\documentclass[runningheads]{llncs}
\usepackage[dvipsnames]{xcolor}

\usepackage{graphicx}
\usepackage{tikz}
\usepackage{comment}
\usepackage{amsmath,amssymb} 
\usepackage{color}

\usepackage[accsupp]{axessibility}  

\usepackage{booktabs}
\usepackage{algorithm}
\usepackage{algorithmic}
\usepackage{amsmath}
\usepackage{bbm}
\usepackage{bm}
\usepackage{color}

\begin{document}

\pagestyle{headings}
\mainmatter
\def\ECCVSubNumber{192}  

\title{Few-shot Single-view 3D Reconstruction with
   Memory Prior Contrastive Network} 

\titlerunning{Few-shot 3D Reconstruction with
   Memory Prior Contrastive Network}
%
\author{Zhen Xing\inst{1} \and
Yijiang Chen\inst{1}  \and
Zhixin Ling\inst{1}   \and \\
Xiangdong Zhou \inst{1}  \and
Yu Xiang \inst{2} }
\authorrunning{Zhen Xing et al.}

\institute{School of Computer Science, Fudan University, Shanghai, China, 200433
\\
\and
The University of Texas at Dallas \hspace{6px}
}
\maketitle

\begin{abstract}

3D reconstruction of novel categories based on few-shot learning is appealing in real-world applications and attracts increasing research interests. Previous approaches mainly focus on how to design shape prior models for different categories. Their performance on unseen categories is not very competitive. In this paper, we present a Memory Prior Contrastive Network (MPCN) that can store shape prior knowledge in a few-shot learning based 3D reconstruction framework. With the shape memory, a multi-head attention module is proposed to capture different parts of a candidate shape prior and fuse these parts together to guide 3D reconstruction of novel categories. Besides, we introduce a 3D-aware contrastive learning method, which can not only complement the retrieval accuracy of memory network, but also better organize image features for downstream tasks. Compared with previous few-shot 3D reconstruction methods, MPCN can handle the inter-class variability without category annotations. Experimental results on a benchmark synthetic dataset and the Pascal3D+ real-world dataset show that our model outperforms the current state-of-the-art methods significantly.

\end{abstract}

\section{Introduction}

\label{sec:intro}
\begin{figure}[t]
\centering
\includegraphics[width=0.9\columnwidth]{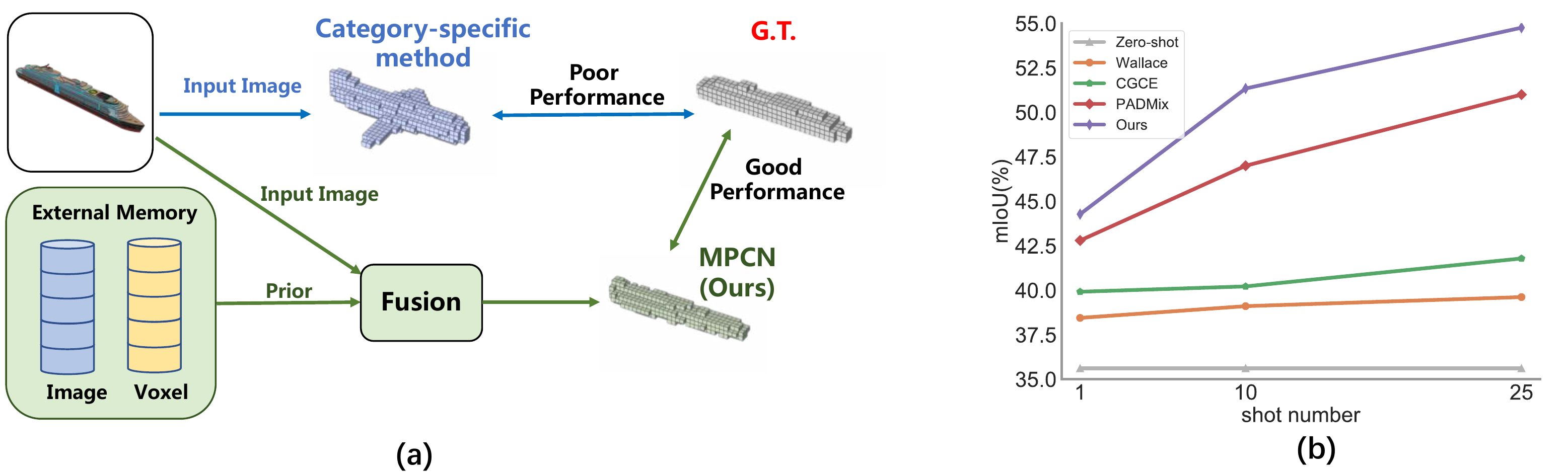} 
\caption{(a) Novel category 3D reconstruction results on category-specific method and our method combination of prior. (b) The mIoU(\%)  of current methods against the number of shots. Our MPCN outperforms the SOTA approaches with different shot.}
\label{fig01}
\end{figure}

Reconstructing 3D shapes from RGB images is valuable in many real-world applications such as autonomous driving, virtual reality, CAD and robotics. Traditional methods for 3D reconstruction such as Structure From Motion (SFM)~\cite{sfm} and Simultaneous Localization and Mapping (SLAM) \cite{slam} often require significant efforts in data acquisition. For example, a large number of images need to be captured and the camera parameters need to be calibrated, which limit the applications of these traditional methods. 

In recent years, 3D reconstruction from single image based on deep neural networks  attracts great research interests. However, previous methods of single-view 3D reconstruction are mostly category-specific \cite{pix2vox,3dr2n2}. Therefore, they only perform well in the specific training categories. These methods also require a large number of labeled training images, which is time consuming and costly to obtain.

Notably, Tatarchenko et al. \cite{tatarchenko2019singlerethinking} shows that single-view 3D reconstruction of specific categories is closely related to image recognition for shape matching. Several simple image retrieval baseline methods can even outperform state-of-the-art 3D reconstruction methods.

In this paper, we propose a novel category-agnostic model for single-view 3D reconstruction in the few-shot learning settings. In our method, the network parameters are first optimized on some specific object categories. Then given novel categories, the model can be quickly transferred to 3D reconstruction of these categories based on few-shot learning techniques. 

To the best of our knowledge, there are mainly three previous works focus on unseen category 3D reconstruction.  Wallace et al. \cite{wallace2019few} present  a network for single-view 3D reconstruction with additional shape prior of object categories. However, their shape prior model cannot catch the diversity of shapes within an object category. The authors of Compositional Global Class Embeddings (CGCE)~\cite{cgce}  adopt a solution to optimize the shape prior codebooks for the reconstruction of unseen classes.  Their model depends on finetuning with additional codebooks, which makes the training process complicated and makes the performance unstable. Pose Adaptive Dual Mixup (PADMix)~\cite{padmix} proposes to apply mixup~\cite{zhang2017mixup} at the feature level and introduce pose level information, which reaches a new state-of-the-art performance in this task.

In addition, all the works rely on shape prior of specific categories. As a result, additional category annotation is needed to recognize the category of the input image. Then these methods can construct shape prior according to the category annotation, which is not very suitable for category-agnostic 3D reconstruction with novel categories.

The previous works of exploring shape prior for 3D reconstruction of novel categories are insightful and reasonable. However, there still exits a challenge on how to handle shape variety within a novel category in the context of few-shot learning.  In this paper, we present a novel deep network model with a memory that can store a shape and its corresponding image as a key-value pair for retrieval. When an image of a novel category is inputted to the network, our deep network can select and combine appropriate shapes retrieved from the memory without category annotation to guide a decoder for 3D reconstruction. In order to adaptively combine the stored shapes as a prior for the downstream 3D reconstruction task, a multi-head attention shape prior module is proposed. Fig. \ref{fig01}(a) shows the example on novel watercraft category 3D reconstruction performance between traditionally category-specific method~\cite{3dr2n2} and our method.

Besides, we propose a 3D-aware contrastive loss that pulls together the image features of objects with similar shape and pushes away the image features with different 3D shape in the latent feature space, which helps for both organizing the image feature and improving the retrieval accuracy of memory network. Our 3D-aware contrastive loss takes into account the difference of shape as a weighting term to reduce or stress the positiveness of a pair, regardless of the category or instance, as we aim at a category-agnostic 3D reconstruction network.

\noindent\textbf{In summary, our contributions are as follows}:
We propose a novel Memory Prior Contrastive Network (MPCN) that can retrieve shape prior as an intermediate representation to help the neural network to infer the shape of novel objects without any category annotations.

Our multi-head attention prior module can automatically learn the association between retrieved prior and pay attention to different part of shape prior. It can not only provide prior information between object categories, but also represent the differences within objects in the same category.

The network with both reconstruction loss and a contrastive loss works together for better result. Our improved 3D-aware contrastive loss takes into account of the difference of positive samples, which is more suitable for supervised 3D tasks.

Experimental results on ShapeNet \cite{chang2015shapenet} dataset show that our method greatly improves the state-of-the-art methods on the two mainstream evaluation metrics using Intersection over Union and F-score. The reconstruction results on the real-world Pascal3D+ \cite{xiang2014beyond} dataset also demonstrate the effectiveness of our method quantitatively and qualitatively.

\section{Related Work}

\noindent\textbf{Deep Learning 3D reconstruction}
Recently, Convolutional Neural Network (CNN) based single-view 3D reconstruction methods become more and more popular. Using voxels to represent a 3D shape is suitable for 3D CNNs. In the early work 3D Recurrent Reconstruction Neural Network (3D-R2N2) \cite{3dr2n2}, the encoder with a Recurrent Neural Network (RNN) structure is used to predict 3D voxels by a 3D decoder. A follow-up work, 3D-VAE-GAN \cite{vae-gan}, explores the generation ability of Variational Autoencoders (VAEs) and Generative Adversarial Networks (GANs) to infer 3D shapes. Marrnet \cite{marrnet} and ShapeHD \cite{shapehd} predict the 2.5D information such as depth, silhouette and surface normal of an input RGB image, and then use these intermediate information to reconstruct the 3D object. OGN\cite{octree} and Matryoshka\cite{matryoshka} utilize octrees or nested shape layers to represent high resolution 3D volume. Pix2Vox \cite{pix2vox} and Pix2Vox++ \cite{pix2vox++} mainly improve the fusion of multi-view 3D reconstruction. Mem3D\cite{mem3d} introduces external memory for category-specific 3D reconstruction. However, its performance relies on a large number of samples saved during the training process. More recently, SSP3D~\cite{ssp3d} propose a semi-supervised setting for 3D Reconstruction. In addition to voxels, 3D shapes can also be represented by point clouds \cite{fan2017point,lin2021single,wang2019mvpnet}, meshes \cite{pixel2mesh,pixel2mesh++} and signed distance fields \cite{xu2019disn,occupancy}.

\noindent\textbf{Few-shot Learning}
Few-shot learning models can be roughly classified into two categories: metric-based methods and meta-based methods. Metric-based methods mainly utilize Siamese networks \cite{koch2015siamese}, match networks \cite{vinyals2016matching,choi2018structured} or prototype networks \cite{gao2019hybrid} to model the distance distribution between samples such that similar samples are closer to each other and heterogeneous samples are far away from each other. Meta-based methods \cite{ravichandran2019few}\cite{ramalho2019adaptive} and meta-gradient based methods \cite{chen2020knowledge,jeong2021few,garcia2017few} are teaching models by using few unseen samples to quickly update the model parameters in order to achieve generalization.

\noindent\textbf{Few-shot 3D Reconstruction}
Wallace et al.\cite{wallace2019few} introduce the first method for single-view 3D reconstruction in the few-shot settings. They propose to combine category-specific shape priors with an input image to guide the 3D reconstruction process. However, in their work, a random shape or a calculated average shape is selected for each category as the prior information, which cannot account for shape diversity among objects in a category. In addition, the method does not explicitly learn the inter-class concepts. Compositional Global Class Embeddings (CGCE)~\cite{cgce} adopts a solution to quickly optimize codebooks for 3D reconstruction of novel categories. Before testing on a novel category, the parameters of other modules are fixed. Only the weight vector of codebooks are optimized with a few support samples from the novel category. Therefore, given a new category, CGCE needs to add a new codebook vector for this category and finetune the weight parameters, which makes the whole process complicated and inefficient. Pose Adaptive Dual Mixup(PADMix)~\cite{padmix} proposes a pose adaptive procedure and a three-stage training method with mixup~\cite{zhang2017mixup}. It tries to solve the pose-invariance issue by an autoencoder but its shape prior module is similar to Wallace~\cite{wallace2019few} and has the drawbacks of complicated three-stage training strategy.

\section{Method}
\begin{figure*}[t]
\centering
\includegraphics[width=0.85\textwidth]{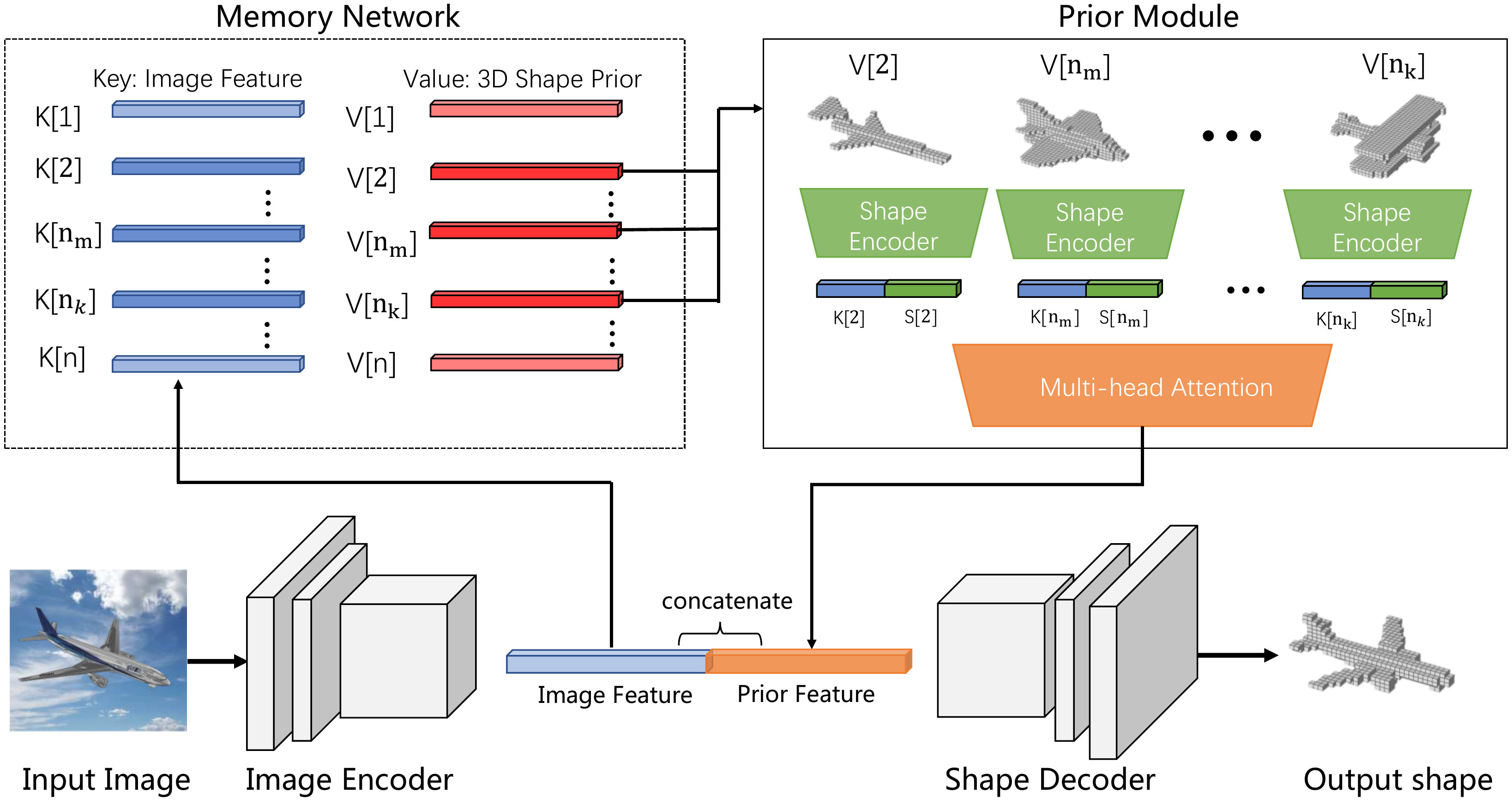}
\caption{An overview of the proposed MPCN. In the training stage, we only use base categories to train the model, and insert memory slots into the memory network as alternative prior according to the rules we set. In the test phase, the memory network saves few shot of support set of the novel category to reconstruct 3D volumes of the query set.}
\label{fig02}
\end{figure*}
Our aim is to design a category-agnostic model, which can achieve superior generalization ability of single-view 3D reconstruction for novel categories with limited support data. Suppose there is a base 3D dataset, defined as $D_b=\{(I_i,V_i)\}_{i=1}^{K}$, where $I_i$ and $V_i$ denote the $i$th image and its corresponding 3D shape represented using voxels, respectively. $K$ is the number of image and voxel pairs in the dataset. We denote the categories in the base dataset  $D_b$ as base categories. Let $D_s=\{(I_i,V_i)\}_{i=1}^{M}$ be another dataset of (image, voxel) pairs with $M$ examples. The categories in $D_s$ are defined as the novel categories, which are different from those in $D_b$. $D_s$ is called the support set, where $M \ll K$.  Meanwhile, there is a large test or query set $D_q=\{(I_i,V_i)\}_{i=1}^{N}$ with $N$ examples. Examples in $D_q$ and $D_s$ are all within the novel categories. Note that we only use $D_b$ and $D_s$ for training. The support set $D_s$ can be used as prior information. We hope that the model can be designed to be category-agnostic and achieve good performance in the query set $D_q$.

\subsection{Memory Network} 

In most previous works on single-view 3D reconstruction, the shape prior information is learned from the model parameters, which leads to category collapse when transferring to novel categories. As mentioned in~\cite{tatarchenko2019singlerethinking}, such kind of model makes the problem degenerated into a classification task. To alleviate this issue, directly using 3D shapes as priors is an intuitive and effective way. As shown in Figure \ref{fig02}, we adopt an explicit key-value memory network to store and calculate shape priors. In the training and testing stages, the CNN features of the input image is extracted by a 2D encoder. Then a retrieval task is performed, where the keys of the samples stored in the memory network are compared to the query vector and the corresponding Top-k retrieved shapes are sent to the prior module for generating prior features. Specifically, as shown in Eq.~\eqref{eq:predict}, the input image $I_q$ is first encoded by a 2D encoder $E_{\text{2D}}$, then the image features and shape prior features are concatenated. Finally, the 3D shape is inferred by a 3D decoder $D_{\text{3D}}$.
\begin{equation} 
\label{eq:predict}
pr=D_{\text{3D}}(\text{Concatenate}(E_{\text{2D}}(I_q), \text{prior feature})),
\end{equation}
where $pr$ is the final predict volume.

\subsubsection{Memory Store} The external memory module is a database of experiences. Each column in the memory represents one data distribution. In MPCN, two columns of structures in the form of key-value are stored. A key is a deep feature vector of an image, and its value is the corresponding 3D shape represented using voxels. Each memory slot is composed of $[\text{image feature}, \text{voxel}]$, and the memory module database can be defined as $\mathcal{M}=\{(I_i,V_i)\}_{i=1}^{m}$, where $m$ represents the size of the external memory module. We use a simple but effective memory storage strategy to store data in a limited memory size. During training stage, when generating a target shape with MPCN, we calculate the distance $d(pr,gt)$ between all the samples' prediction and target shape of a batch as in Eq.~\eqref{eq:dist}:
\begin{equation} 
\label{eq:dist}
d(pr,gt)=\frac{1}{{r_v}^3}\sum_{i,j,k}{(pr_{(i,j,k)}-gt_{(i,j,k)})}^2,
\end{equation}
where $r_v$ is the resolution of 3D volumes, $gt$ is the ground truth volume. For a sample $(I_k,V_k)$ , if $d(pr,gt)$ is greater than a specified threshold $\delta $, we consider that the current network parameters and the prior have poor reconstruction performance on this shape. So we insert $(I_k,V_k)$ into the external memory module and store it as a memory slot in order to guide the reconstruction of similar shapes in the future. We maintain an external memory module similar to the memory bank(queue). When the memory is full, the memory slot initially added to the queue will be replaced by later one. This makes sense because the later image features are updated with iteration of the model training.

\subsubsection{Memory Reader} Each row of the external memory database represents a memory slot. The retrieval of the memory module is based on a k-nearest neighbor algorithm. When comparing the CNN features of the current query image and the image features of all slots in the memory network, we use the Euclidean metric to measure the differences. In order to obtain the distance between the query matrix and the key matrix of the memory network conveniently, we use the effective distance matrix computation to calculate the matrix of Euclidean distance as shown in Eq.~\eqref{eq:matrix}:
\begin{equation} \label{eq:matrix}
\text{Distance}=\Vert{Q}\Vert+\Vert{K}\Vert-2*QK^T,
\end{equation}
where $Q\in\mathbbm{R}^{b\times 2048}$ is query matrix, $K\in\mathbbm{R}^{m\times 2048}$ is memory-key matrix, $b$ is the batch size, and $m$ is the memory size.

After calculating the distance, we select the nearest $k$ retrieval results as prior information, which is defined as $R=\{(I_i,V_i)\}_{i=1}^{k}$. $\{I_i\}_{i=1}^{k}$ represents $k$ retrieved image features, $\{V_i\}_{i=1}^{k}$ represents $k$ retrieved voxels. When the first batch is searched, the memory will be empty. However, we will take out $k$ all-zero tensors, which increases the robustness of the model to some extent. Even without the shape prior as a guide, our model should reconstruct the 3D model according to the 2D features of the image.

\subsection{Prior Module}

\begin{figure*}[t]
\centering
\includegraphics[width=0.88\textwidth]{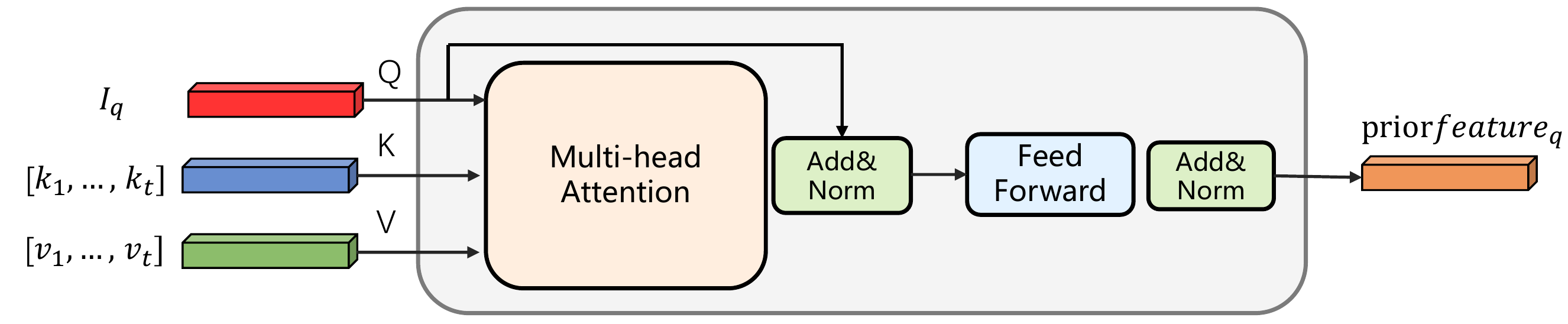} 
\caption{An overview of the proposed Memory Prior Module.}
\label{fig3}
\end{figure*}

The prior module can first obtain the set $\{(I_i,V_i)\}_{i=1}^{k}$ retrieved by the external memory module from the previous step. Note that shape volume is the original voxel saved at this stage, and its size is $32^3$. So the model needs to extract shape features by a 3D shape encoder before the downstream processing.
\begin{equation}
k_i=I_i,v_i=\text{Encoder3D}(V_i),
\end{equation}
\begin{equation}
\label{eq:5}
Q = I_q W_q, K = k_i W_k, V = v_i W_v,
\end{equation}
\begin{equation}
\label{eq:6}
e_q=Q+\text{LayerNorm}(\text{MHA}(Q,K,V)),
\end{equation}
\begin{equation}
\label{eq:7}
{prior\ feature}_q=e_q+\text{LayerNorm}(\text{FFN}(e_q)),
\end{equation}

In previous works, only 3D voxels are regarded as the prior features. In contrast, as shown in Figure \ref{fig3}, we use the attention based architecture to extract shape prior features by exploring the association between image features and 3D shape. Concretely, we take the query image feature $I_q$ as the query, the retrieved image feature $\{I_i\}_{i=1}^{k}$ as the key, and its corresponding 3D shape feature $\{v_i\}_{i=1}^{k}$ as the value. As shown in Eq.~\eqref{eq:5}, we first use three separate linear layers parameterized by $W_q$, $W_k$ and $W_v$ to extract query, key, value embedding Q, K and V.

Then the embeddings are forwarded to the multi-head attention(MHA)\cite{vaswani2017attention} and layer normalization(LayerNorm) module\cite{ba2016layer} to perform cross-attention between the query and every key. The output of the attention is fused to the original input query embedding to get enhanced feature $e_q$. Afterward, the obtained features $e_q$ are sent into feed-forward network(FFN) and layer normalization(LayerNorm). The output ${prior feature}_q$ is obtained by adding up the feed-forward module output with residual connection as in Eq.~\eqref{eq:6} and Eq.~\eqref{eq:7}.

\subsection{3D-Aware Contrastive Learning Method}

\begin{figure}[t]
\centering
\includegraphics[width=0.7\columnwidth]{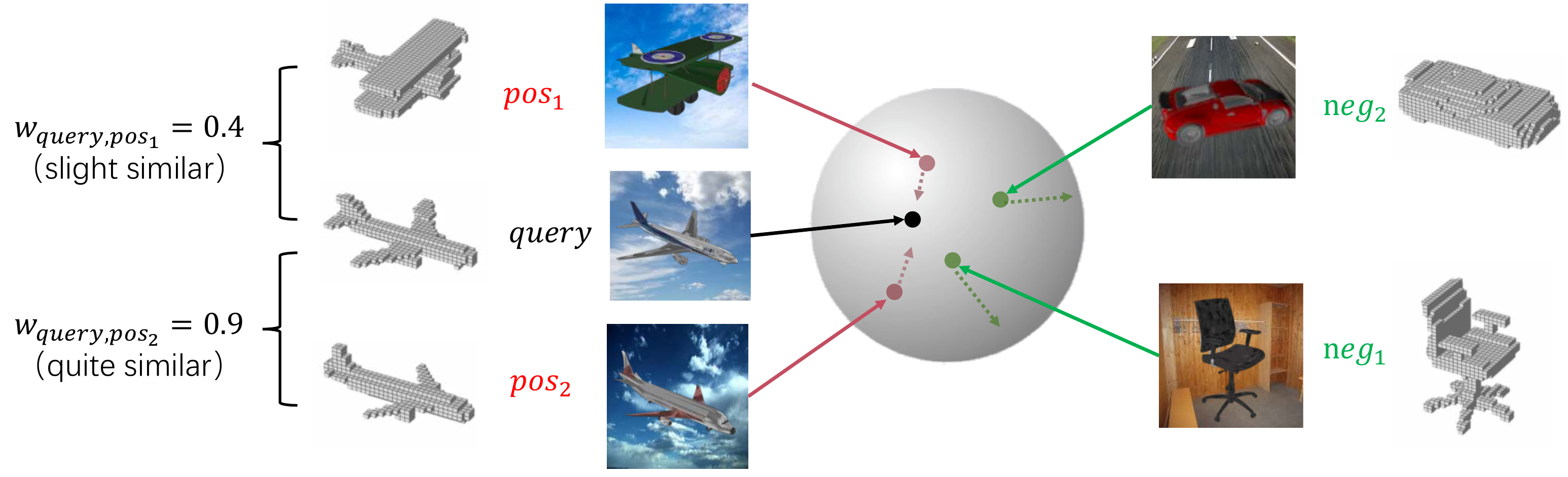} 
\caption{An example of 3D-Aware Contrastive Loss, which pulls together the positive samples with similar 3D shape(e.g., \textcolor{red}{$pos_1$} and \textcolor{red}{$pos_2$}) by different $weight$, and pushes apart the negative samples with different 3D shape(e.g., \textcolor{green}{ $neg_1$} and \textcolor{green}{$neg_2$}).}
\label{fig04}
\end{figure}

We believe that Memory Prior Network can work effectively mainly based on the accuracy of 2D image embedding retrieval. In order to improve the retrieved prior accuracy provided by memory network, previous work generally used triple loss\cite{hermans2017defense} to optimize encoder, which is effective for simple classification problems \cite{kaiser2017learning} \cite{zhu2018compound}. However, for 3D reconstruction task, the triplets need to construct positive and negative samples according to the threshold of specific shape difference, which is an empirical and troublesome step \cite{mem3d}. 

Recently, contrastive learning method \cite{chen2020simple}\cite{he2020momentum} train an image encoder maximizing the similarity of different transformations of the same sample and minimizing the similarity with other samples. The proposed loss ${\cal{L}}_{infoNCE}$ \cite{he2020momentum} achieves great success in self-supervised representation learning. However, their success depends on the large batch size and memory occupation, and taking all of samples in the same batch as negative pairs may be wrong for supervised tasks.
In addition, supervised contrastive loss\cite{khosla2020supervised} tries to solve this problem but it mainly focuses on simple classification problem. 

We hope to design a loss that can adaptively pull together the image embedding pairs with similar 3D shape and push away the image embeddings with different 3D shapes. Therefore, as shown in Fig.~\ref{fig04}. we introduce an improved 3D-aware contrastive loss, which considers the positivity of positive pairs. Concretely, for each image pair $(q, k)$, we calculate the distance between the associated 3D shape $d(V_q, V_k) \in [0,1]$, we take $(q,k)$ as a positive pair if $d(V_q, V_k) < \delta$, then a weight is calculated by $w_{q,k}=(1-d(V_q,V_k)\times \gamma)$, which is considered as the important weight of the positive pair in our 3D-aware contrastive loss:

\begin{equation}
{\cal{L}}_{3DNCE}=-log\frac{\sum_{p \in [1..M]} w_{q,p}\cdot exp(f_q \cdot f_{p}/\tau) }{M \cdot \sum_{k \in [1..N]} exp(f_q \cdot f_k/ \tau).}
\end{equation}

Where $d(V_q,V_p)$ is the same as Eq~\eqref{eq:dist}, $q$ is a query image, $p$ is the positive samples of $q$ according to $d(V_q,V_p) < \delta$, $\sum_{k \in [1..N]}$ mean the samples in the same batch with $q$, and $f$ is image encoder. Intuitively, the more similar the 3D shape of two objects $(q, p)$ is, the greater its weight $w_{q,p}$ is, and the closer the image features of the two objects are.

\subsection{Training Procedure in Few-shot settings}
We adopt a two-stage training strategy. In the first stage, we train the initialization model on the base category data $D_b$. In the second stage, we use few-shot novel category samples in support set $D_s$ to finetune the network. Both stage adopt the training method based on episodes as shown in the Algorithm \ref{alg:algorithm1}. At the beginning of each epoch, all slots of the memory are cleared to ensure that the new round of training can re-determine which samples are inserted into the memory module according to our memory store strategy. For test phase, we first insert samples in support set $D_s$ to the memory module as candidate prior information according to the few-shot settings. Then it follows the same steps as training stage to predict 3D shape in query set $D_q$. Finally, the reconstruction results are evaluated by evaluation metric.

\begin{algorithm}[tb]

\algsetup{linenosize=\scriptsize} \small
\caption{Training algorithm}
\label{alg:algorithm1}
\begin{algorithmic}[1] 
\FOR{epoch in epochs}
\STATE flush memory slots
\FOR{batch\_idx in range(max\_episode):}
\STATE Load query images, target shape from train set $D_b$
\STATE embed2d $=$ \textit{Encoder2d}(query image)
\STATE Key, Value, Distance $=$ \textit{Top-K}(embed2d)
\STATE embedPrior $=$ \textit{Prior}(embed2d, Key, Value, Distance)
\STATE embed $=$ \textit{concatenate}(Embed2d, embedPrior)
\STATE predict $=$ \textit{Decoder3d}(embed)
\STATE d $=$ \textit{computeDis}(predict, target shape)
\IF{$d >\delta$:}
\STATE insert(image, voxel) to external memory
\ENDIF
\STATE Train on predict and target with backprop
\ENDFOR
\ENDFOR
\end{algorithmic}

\end{algorithm}

\subsection{Architecture} 
\noindent\textbf{Image Encoder} The 2D encoder shares the same ResNet backbone \cite{he2016deep} as that of CGCE \cite{cgce}. Then our model follows with three layers of convolution layer, batch normalization layer and Relu layer. The convolution kernel size of the three convolution layers is $3^2$ with padding 1. The last two convolution layers are then followed by a max pooling layer. The kernel size of the pooling layer is $3^2$ and $2^2$, respectively. The output channels of the three convolution layers are 512, 256 and 128, respectively. 

\noindent\textbf{Prior Module} The 3D shape encoder of the prior module includes four convolution layers and two max-pooling layers. Each layer has a LeakyRelu activation layer, and the convolution kernel sizes are $5^3$, $3^3$, $3^3$ and $3^3$, respectively. The output Q,K,V embedding dimension of the Linear layer is 2048. The size of key and value of the attention module are both 2048. This module has 2 heads in attention blocks. Finally, the prior feature dimension is 2048.

\noindent\textbf{Shape Decoder} There are five 3D deconvolution layers in this module. The convolution kernel size is $4^3$, stripe is $2^3$, and padding is 1. The first four convolutions are followed by batch normalization and Relu, and the last is sigmoid function. The output channels of the five convolution layers are 256, 128, 32, 8 and 1, respectively. The final output is a voxel representation with size $32^3$.

\subsection{Loss Function}
\subsubsection{Reconstruction Loss}
For the 3D reconstructions network, both the reconstruction prediction and the ground truth are based on voxel. We follow previous works \cite{wallace2019few,cgce,pix2vox,pix2vox++} that adopt the binary cross entropy loss as our loss function:
\begin{equation}
{\cal{L}}_{rec}=\frac{1}{r_{v}^{3}}\sum_{i=1}^{r_{v}^{3}}[{gt}_i\log({pr}_i)+(1-{gt}_i)\log(1-{pr}_i)],
\end{equation}
where $r_v$ represents the resolution of the voxel space, $pr$ and $gt$ represent the predict and the ground truth volume.

\subsubsection{Total Loss} The MPCN is trained end-to-end with the reconstruction loss and 3D-aware contrastive loss together as following:
\begin{equation}
{\cal{L}}_{total}= {\cal{L}}_{rec} + \lambda {\cal{L}}_{3DNCE}
\end{equation}
where $\lambda$ is hyperparameter, which is set to 0.001 in this work.

\begin{figure}[t]
\centering
\includegraphics[width=0.85\columnwidth]{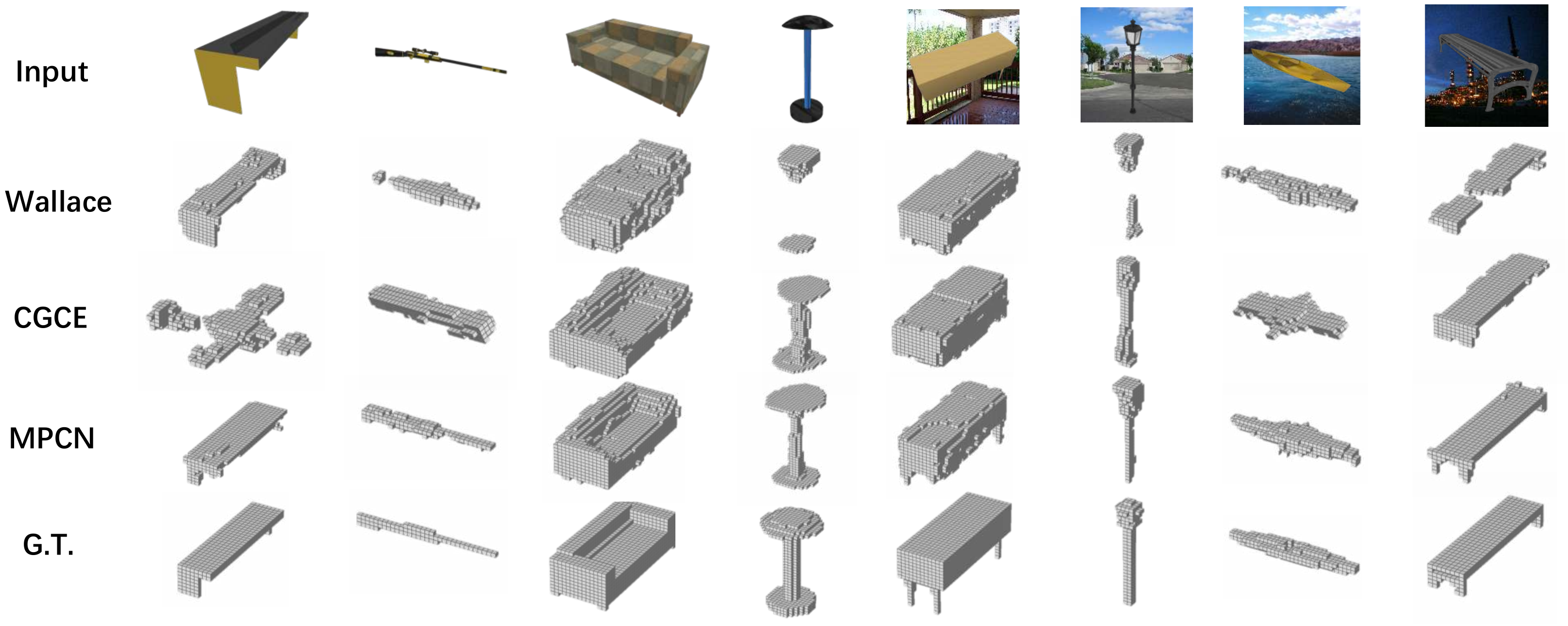} 
\caption{Examples of single-view 3D Reconstruction on novel category of ShapeNet with shot-10. We show examples with clean background and with random background.}
\label{fig05}
\end{figure}

\section{Experiment}
\subsection{Experimental setup}
\subsubsection{Dataset} We experiment with the \textbf{ShapeNet} dataset \cite{chang2015shapenet}. The setting of this dataset is the same as \cite{wallace2019few}. Seven of 13 categories are designated as the base classes that are used for training: airplanes, cars, chairs, displays, phones, speakers and tables. The other categories are set as novel classes for testing. To fairly compare with the previous works, we use the same dataset split as in \cite{wallace2019few} and CGCE \cite{cgce}. The datasets are provided by \cite{3dr2n2} which are composed with $137\times137$ RGB images and $32\times32\times32$ voxelized representations.
 \textbf{Pascal3D+} \cite{xiang2014beyond} dataset has 12 different categories. It provides approximate 3D annotations for Pascal VOC2012 and ImageNet \cite{xiang2014beyond}. Each category has about 10 CAD models, which are generally not used for training directly. We finetune our MPCN with 8 categories of Pascal3D+, and test it on four categories: bicycle, motorbike, bottle and train. For fair comparison, we use Binvox \cite{nooruddin2003simplification} tool to render the voxel representation from the CAD model, and the voxel resolution is also $32\times32\times32$.

\begin{table*}
\caption{Comparison of single-view 3D object reconstruction on novel class of ShapeNet at $32^3$ resolution with different available shot.We report the \textbf{mean IoU} per category. The best number for each category is highlighted in bold.} 

\centering
\scalebox{0.75}{
\begin{tabular}{@{}c|c|cccc|cccc|cccc@{}}
\toprule
           & 0-shot & \multicolumn{4}{c|}{1-shot}                     & \multicolumn{4}{c|}{10-shot}                             & \multicolumn{4}{c}{25-shot}             \\ \midrule
category   & B0     & Wallace & CGCE & PADMix        & MPCN          & Wallace & CGCE          & PADMix        & MPCN          & Wallace & CGCE & PADMix & MPCN          \\ \hline
cabinet    & 0.69   & 0.69    & 0.71 & 0.67          & \textbf{0.72} & 0.69    & \textbf{0.71} & 0.66          & 0.68          & 0.69    & 0.71 & 0.68   & \textbf{0.74} \\
sofa       & 0.52   & 0.54    & 0.54 & 0.54          & \textbf{0.57} & 0.54    & 0.54          & 0.57          & \textbf{0.60} & 0.54    & 0.55 & 0.59   & \textbf{0.65} \\
bench      & 0.37   & 0.37    & 0.37 & 0.37          & \textbf{0.39} & 0.36    & 0.37          & \textbf{0.41} & \textbf{0.41} & 0.36    & 0.38 & 0.42   & \textbf{0.45} \\
watercraft & 0.28   & 0.33    & 0.39 & \textbf{0.41} & \textbf{0.41} & 0.36    & 0.41          & 0.46          & \textbf{0.54} & 0.37    & 0.43 & 0.52   & \textbf{0.55} \\
lamp       & 0.19   & 0.20    & 0.20 & \textbf{0.29} & \textbf{0.29} & 0.19    & 0.20          & 0.31          & \textbf{0.32} & 0.19    & 0.20 & 0.32   & \textbf{0.37} \\
firearm    & 0.13   & 0.21    & 0.23 & \textbf{0.31} & 0.24          & 0.24    & 0.23          & 0.39          & \textbf{0.52} & 0.26    & 0.28 & 0.50   & \textbf{0.52} \\ \hline
mean       & 0.36   & 0.38    & 0.40 & 0.43          & \textbf{0.44} & 0.40     & 0.41          & 0.47          & \textbf{0.51} & 0.41    & 0.43 & 0.51   & \textbf{0.54} \\ \bottomrule
\end{tabular}

}

\label{table1}
\end{table*}

\noindent\textbf{Evaluation Metrics}
For fair comparison, we follow previous work \cite{wallace2019few}\cite{cgce}\cite{pix2vox} using \textbf{Intersection over Union (IoU)} as the evaluation metrics. The IoU is defined as following:

\begin{equation}
\text{IoU}=\frac{\sum_{i,j,k}{\cal{I}}(\hat{p}_{(i,j,k)}>t){\cal{I}}(p_{(i,j,k)})}{\sum_{i,j,k}{\cal{I}}[{\cal{I}}(\hat{p}_{(i,j,k)}>t)+{\cal{I}}(p_{(i,j,k)})]},
\end{equation}
where $\hat{p}_{(i,j,k)} $ and $p_{(i,j,k)}$represent the predicted possibility and the value of ground truth at point $(i,j,k)$, respectively. $\cal{I}$ is the function that is one when the requirements are satisfied. $t$ represents a threshold of this point, which is set to 0.3 in our experiment.

\subsubsection{Implementation details}
We used $224\times 224$ RGB images as input to train the model with a batch size of 16. Our MPCN is implemented in PyTorch \cite{paszke2019pytorch} and trained by the Adam optimizer \cite{kingma2014adam}. We set the learning rate as $1e-4$. The $\delta$ and $\gamma$ are set to 0.1 and 10. The $k$ of the retrieval samples Top-k is 5, the $\tau$ is set 0.1. The memory size $m$ is set to 4000 in the training stage, and only 200 in the testing stage.

\subsubsection{Baseline}
We compare our proposed MPCN with three state-of-the-art methods: Wallace\cite{wallace2019few}, CGCE \cite{cgce} and PADMix\cite{padmix}. We also  follows the zero-shot lower baseline in CGCE \cite{cgce}. Zero-shot refers to the result of training on the base categories with only single-image and testing directly on the novel class without any prior. Image-Finetune method refers to training on the base categories and finetuning the full network with few available novel categories samples.

\subsection{Results on ShapeNet dataset}
We compare with the state-of-the-art methods on the ShapeNet novel categories. Table \ref{table1} shows the $\text{IoUs}$ of our MPCN and other methods. Results in Table \ref{table1} from other models are taken from \cite{padmix}. For few-shot settings, we follow the evaluation in CGCE \cite{cgce} and PADMix\cite{padmix} shown the results in the settings of 1-shot, 10-shot and 25-shot. It can be seen that our method is much better than the zero-shot baseline respectively, and greatly outperforms SOTA's methods. Experimental results show that MPCN has great advantages when the shot number increases, mainly because it can retrieve prior information more related to the target shape. Even in the results of 1-shot, there are some improvements, mainly because our model can select the most appropriate prior of shapes as well as image features, and use the differences of other shapes to exclude other impossible shapes. Our MPCN results are shown in Figure \ref{fig04}. It can be seen that our model can obtain satisfactory reconstruction results for novel categories than any other SOTA methods even when the angles of input images are different.

\begin{table}

\caption{Comparison of single-view 3D object reconstruction on Pascal3D+ at $32^3$resolution. We report both the mean IoU of every novel category. The best number is highlighted in bold.} 
\centering

\scalebox{0.9}{
\centering
\begin{tabular}{@{}c|cccc|c@{}}
\toprule
               & bicycle       & motorbike     & train         & bottle        & mean            \\ \midrule
Zero-shot      & 0.11          & 0.27          & 0.35          & 0.10          & 0.2074          \\ \hline
Image-Finetune & 0.20          & 0.28          & 0.35          & 0.32          & 0.2943          \\
Wallace~\cite{wallace2019few} & 0.21          & 0.29          & \textbf{0.40} & 0.43          & 0.3324           \\
CGCE~\cite{cgce}& 0.23          & 0.33          & 0.37          & 0.35          & 0.3223          \\ \hline
MPCN(ours)     & \textbf{0.28} & \textbf{0.39} & 0.37          & \textbf{0.46} & \textbf{0.3748} \\ \bottomrule
\end{tabular}

}

\label{table3} 
\end{table}

\subsection{Results on Real-world dataset}
In order to compare with Wallace et al. \cite{wallace2019few} further, we also conducted experiments on the real-world Pascal3D+ dataset \cite{xiang2014beyond}. Firstly, the model is pre-trained on all 13 categories of the ShapeNet dataset \cite{chang2015shapenet}. Then the model is finetuned with Pascal3D+ base category dataset, and the final test set is selected from the four novel categories. Note that the experiment is set of 10-shot. The experimental results show that our method outperform zero-shot baseline by 16.74\%, also it is the best compared with SOTA methods. Especially for bicycle and motorbike categories with large shape difference and variability, our model perform best. But for the category with subtle shape difference (\emph{e.g.}, train), the reconstruction results tend to align to the average of prior shape, so Wallace~\cite{wallace2019few} shows marginal improvement over ours. But note that our MPCN selects prior information by memory network automatically, while Wallace et al.~\cite{wallace2019few} need the category annotations of the input images and choose shape priors manually. The experimental results are shown in the Table \ref{table3}. 

\begin{table}
\centering
\caption{The effectiveness of the different modules and losses. All the results are tested on novel categories of ShapeNet and Pascal3D+ with 5-shot. We report the mean IoU(\%) of novel categories. The best number is highlighted in bold.} 

\scalebox{0.8}{
\begin{tabular}{@{}c|ccccccc|c|c@{}}
\toprule
               & MHA & LSTM & Random & Average & Finetune & ${\cal{L}}_{3DNCE}$ & ${\cal{L}}_{infoNCE}$  & ShapeNet  & Pascal3D+ \\ \midrule
Zero-shot      &     &      &        &         &          &           &          & 35.68 & 20.74     \\ 
ONN(5)        & & & & & & & &40.90 & 42.50 \\ \hline
Image-Finetune &     &      &        &         & \checkmark        &           &         & 40.82 & 27.49     \\
MPCN-Top1      & \checkmark   &      &        &         & \checkmark        & \checkmark         &         & 39.95 & 31.65     \\ \hline
w LSTM           &     & \checkmark    &        &         &          & \checkmark         &         & 40.52 & 25.63     \\
w Random prior    &     &      & \checkmark      &         &          & \checkmark         &         & 35.52 & 22.05     \\
w average   &     &      &        & \checkmark       &          & \checkmark         &         & 41.32 & 29.85     \\ \hline
w ${\cal{L}}_{infoNCE}$       & \checkmark   &      &        &         &          &           & \checkmark    &    42.98 & 27.98     \\

w/o ${\cal{L}}_{3DNCE}$       & \checkmark   &      &        &         &          &           &         & 43.05 & 28.76     \\
w/o finetune   & \checkmark   &      &        &         &           & \checkmark         &         & 45.82 & 30.32     \\ \hline
ours           & \checkmark   &      &        &         & \checkmark        & \checkmark         &         & \textbf{47.53} & \textbf{33.54}     \\ \bottomrule
\end{tabular}

}
\label{table3} 
\end{table}

\section{Ablation Study}
In this part, we evaluate the effectiveness of proposed modules and the impact of different losses. We choose the setup of 5-shot on both ShapeNet and Pascal3D+ dataset for comparative experiment if not mentioned elsewhere.

\noindent\textbf{Retrieval or Reconstruction}  In order to prove that our method is superior to the retrieval method, we just take the highest similarity retrieved shape as the target shape. That is the result of MPCN-Top1 in Table \ref{table3}. In addition, Image-Finetune method is also shown for comparison. The results in Table~\ref{table3}  shows that our MPCN outperforms any upper retrieval or finetune methods in the few-shot settings.

\noindent\textbf{Analysis of Prior Module} We analyze the prior extraction module based on attention in MPCN. Because previous methods using external memory network adopt LSTM \cite{lstm} in the shape prior fusion stage, we replace the attention part of MPCN with LSTM (w LSTM) for the purpose of comparison. We also compare the average-fusion (w average) of the retrieved Top-5 object volume features. In addition, the random initialization of prior vectors (w Random prior) is also compared in the experiment. 

\begin{figure}[t]
\centering
\includegraphics[width=0.9\columnwidth]{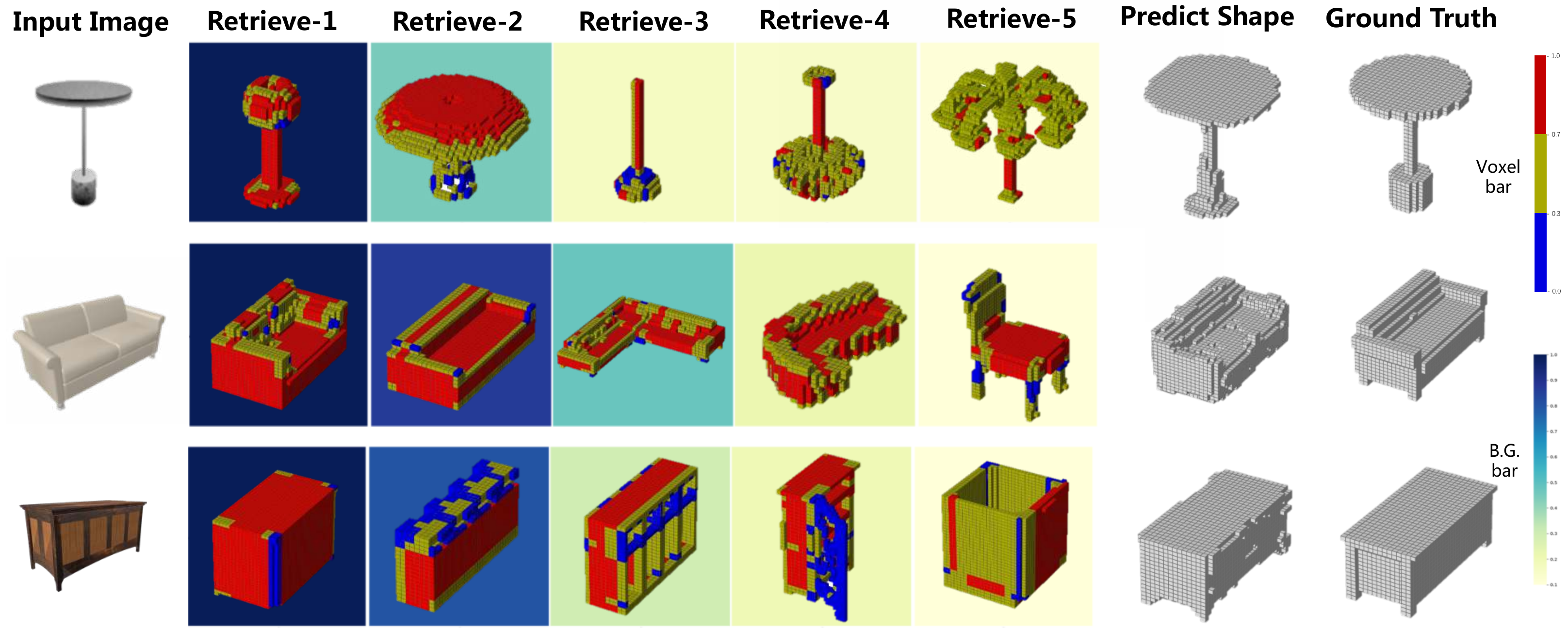} 
\caption{Visualization of features maps from the retrieved 3D volumes and the corresponding reconstructions.}
\label{fig06}
\end{figure}

The experimental results in Table \ref{table3} show that our prior module plays an important role for guiding the reconstruction of 3D objects. The fusion of Top-5 average method and random prior obviously cannot make full use of similar 3D volumes. Our attention module can capture the relevance of different 3D objects better than LSTM, and shows more powerful ability of inferring 3D shapes by using prior information in the few-shot settings. Fig. \ref{fig06} illustrates some shape priors selected by our model.  We demonstrate that the multi-head attention module can adaptively detect the proper parts of the retrieved shapes for 3D reconstruction of novel categories. 

\noindent\textbf{Analysis of Loss and Finetune} In order to further prove the effectiveness of our proposed 3D-aware contrastive loss, we remove ${\cal{L}}_{3DNCE}$ in the training stage, as (MPCN w/o ${\cal{L}}_{3DNCE}$) shown in Table \ref{table3}. Besides, we replace the improved ${\cal{L}}_{3DNCE}$ with the traditional contrastive loss, as shown (MPCN w ${\cal{L}}_{infoNCE}$) in Table \ref{table3}. The results show that our comparison ${\cal{L}}_{3DNCE}$ has a great contribution to the improvement of experimental effect, mainly because it not only improves the retrieval accuracy of memory a prior module, but also makes the intent space of 2D representation more reasonable. In addition, using few-shot novel category samples in support set $D_s$ to finetune the network (w finetune) in the second training stage is also important for the performance. 

\section{Conclusion}
In this paper, we propose a novel category-agnostic model for 3D object reconstruction. Inspired by the novel 3D object recognizing ability by human-beings, we introduce an external memory network to assist in guiding the object to reconstruct the 3D model in few-shot settings. Compared with the existing methods, our method provides an advanced module to select shape priors, and fuses shape priors and image features to reconstruction 3D shapes. In addition, a 3D-aware contrastive method is proposed for encode 2D latent space, which may be used for other supervised tasks of 3D vision. The experimental results show that our MPCN can outperform existing methods on the ShapeNet dataset and the Pascal3D+ dataset under the settings of few-shot learning. 

\noindent \textbf{Acknowledgments:}
This work was supported by the National Key Research and Development Program of China, No.2018YFB1402600.

\bibliographystyle{splncs04}
\bibliography{egbib}
\end{document}